\documentclass[lettersize,journal]{IEEEtran}
\usepackage{amsmath,amsfonts}
\usepackage[linesnumbered,ruled]{algorithm2e}
\usepackage{array}
\usepackage[caption=false,font=normalsize,labelfont=sf,textfont=sf]{subfig}
\usepackage{textcomp}
\usepackage{stfloats}
\usepackage{url}
\usepackage{verbatim}
\usepackage{graphicx}
\usepackage{cite}

\usepackage{booktabs} 
\usepackage{bbding}  
\usepackage{color}
\usepackage[pagebackref=true,breaklinks=true,colorlinks,bookmarks=false,allcolors=blue]{hyperref}
\usepackage[table]{xcolor}
\begin{document}

\title{SAMOFT: Robust Multi-Object Tracking via Region and Flow}


\author{
Yanchao Wang, Dawei Zhang, Chengzhuan Yang, Wei Liu, Minglu Li,~\IEEEmembership{Fellow,~IEEE}, Hua Wang,~\IEEEmembership{Fellow,~IEEE}, Zhonglong Zheng, and Ming-Hsuan Yang,~\IEEEmembership{Fellow,~IEEE}
\thanks{This research was supported by the National Natural Science Foundation of China under Grants No. 62272419 and No. 62402449, and the Major Program of the Natural Science Foundation of Zhejiang Province under Grant No. LD26F020003. \textit{(Yanchao Wang and Dawei Zhang contributed equally to this work.) (Corresponding author: Zhonglong Zheng.)}}
\thanks{Yanchao Wang, Dawei Zhang, Chengzhuan Yang, Minglu Li, and Zhonglong Zheng are with the School of Computer Science and Technology,
Zhejiang Normal University, Jinhua 321004, China  (email: yanchaowang@zjnu.edu.cn; davidzhang@zjnu.edu.cn; czyang@zjnu.edu.cn; mlli@zjnu.edu.cn; zhonglong@zjnu.edu.cn).}
\thanks{Wei Liu is with the School of Automation and Intelligent Sensing, the Institute of Image Processing and Pattern Recognition, and the Institute of Medical Robotics, Shanghai Jiao Tong University, Shanghai 200240, China (e-mail: weiliucv@sjtu.edu.cn).}
\thanks{Hua Wang is with the Institute for Sustainable Industries and Liveable Cities, College of Engineering and Science, Victoria University, Melbourne, VIC 8001, Australia (e-mail: hua.wang@vu.edu.au)}
\thanks{Ming-Hsuan Yang is with the School of Electrical Engineering and Computer Science, University of California at Merced, Merced, CA 95343 USA (e-mail: mhyang@ucmerced.edu).}
}



\maketitle

\begin{abstract}
Multi-object tracking (MOT) is a fundamental task in computer vision that requires continuously tracking multiple targets while maintaining consistent identities across frames.
However, most existing approaches primarily rely on instance-level object features for trajectory association, which often leads to degraded performance under challenging conditions such as object deformation, nonlinear motion, and occlusion. 
In this work, we propose SAMOFT, a robust tracker that leverages pixel-level cues to improve robustness under complex motion scenarios.
Specifically, we introduce a Pixel Motion Matching (PMM) module that integrates the Segment Anything Model (SAM) with dense optical flow to refine Kalman filter-based motion prediction using instantaneous foreground pixel motion. 
To further enhance robustness under unreliable detections, we design a Centroid Distance Matching (CDM) module that performs flexible mask-based centroid matching for low-confidence or partially occluded observations. 
Moreover, a Distribution-Based Correction (DBC) module models long-tailed motion patterns in a training-free manner using historical optical flow statistics and dynamically corrects trajectory states online. 
We also incorporate a Cluster-Aware ReID (CA-ReID) strategy to improve the stability and discriminative power of trajectory appearance features. 
Extensive experiments on the DanceTrack and MOTChallenge benchmarks demonstrate that SAMOFT consistently improves baseline trackers and achieves competitive performance compared with recent state-of-the-art methods, validating the effectiveness of leveraging pixel-level cues for robust multi-object tracking.
\end{abstract}


\section{Introduction}
\IEEEPARstart{M}{ulti-object} tracking (MOT) is a fundamental problem in computer vision and plays an important role in applications such as video surveillance~\cite{milan2016mot16} and motion analysis~\cite{sun2022dancetrack}. 
The long-standing tracking-by-detection paradigm~\cite{bewley2016simple,cao2023observation,yang2024hybrid} decomposes MOT into two subproblems: object detection in each frame and trajectory association across frames. 
With the rapid progress of modern deep object detectors~\cite{ge2021yolox}, detection accuracy has improved substantially, making the association stage the primary factor affecting overall tracking performance. 
In typical tracking-by-detection pipelines, the association stage constructs a cost matrix based on motion and appearance cues and solves the optimal assignment problem using the Hungarian algorithm.
\begin{figure}[t]
\centering
\includegraphics[width=1.\linewidth]{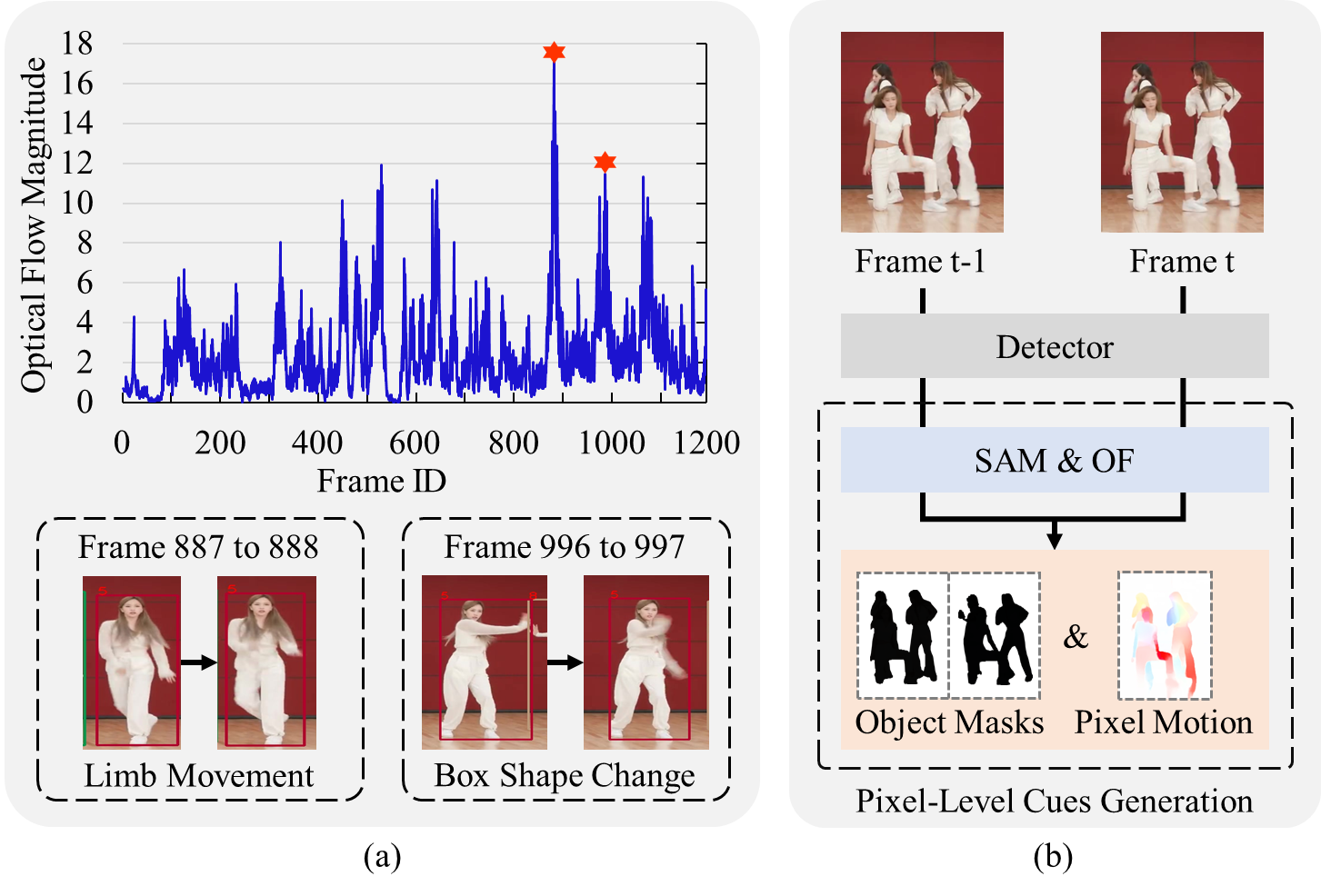} 
\caption{(a) Pixel motion (i.e., optical flow magnitude) effectively captures drastic motion events at local peaks (marked by red stars).
(b) SAMOFT incorporates the Segment Anything Model (SAM) and an optical flow (OF) model to generate object masks and pixel motion, enabling more robust association through pixel-level cues.}
\label{fig: motivation}
\end{figure}

A major line of research under the tracking-by-detection paradigm therefore focuses on designing effective association strategies and cost formulations to better exploit motion and appearance cues. 
Existing studies have explored confidence-based cascaded matching strategies~\cite{zhang2022bytetrack}, incorporated motion-direction information into the association cost~\cite{cao2023observation}, leveraged auxiliary weak cues such as detection confidence and object height~\cite{yang2024hybrid}, and designed adaptive appearance update mechanisms~\cite{maggiolino2023deep}.
However, these explorations remain at the instance level.
Although they achieve strong performance in pedestrian tracking, their reliance on bounding-box geometry and patch-level embeddings limits discriminability under challenging conditions such as nonlinear motion, non-rigid deformation, and occlusion, often leading to suboptimal performance.

To enhance robustness in such challenging scenarios, recent studies have explored integrating multi-modal cues from additional sensors (e.g., depth~\cite{dendorfer2022quo} or thermal infrared~\cite{el2025thermal,zhu2025visible}) or multi-view camera systems~\cite{he2020multi,luo2025omnidirectional} to enable more comprehensive scene modeling.
While effective, these approaches impose additional requirements on data annotation, hardware support, and computational resources for modality fusion, which may limit practical deployment.

In contrast, we focus on the raw image and explore finer-grained pixel-level cues. A natural starting point is dense optical flow, which provides instantaneous motion estimates for every pixel. Unlike instance-level motion modeling based on bounding boxes, pixel-wise motion remains reliable under partial occlusion or bounding-box distortion, where global motion estimation becomes unstable.
Moreover, optical flow captures abrupt local variations when a target deviates from its regular motion pattern. As illustrated in Fig.~\ref{fig: motivation}(a), we analyze temporal pixel-wise motion within an object. The mean optical flow magnitude shows periodic fluctuations aligned with the target's motion cycle, while local peaks correspond to long-tailed events. For example, at frames 888 and 997, sharp increases in flow magnitude coincide with limb movements and bounding-box deformation.
These observations suggest that pixel-level motion cues can anticipate state transitions before global displacement becomes apparent.

However, directly extracting optical flow from bounding boxes introduces contamination from background pixels and nearby overlapping objects, compromising the reliability of pixel-level cues and potentially misleading motion modeling. 
This motivates integrating a segmentation model to isolate true target pixels.
Since MOT datasets typically lack pixel-level annotations, training task-specific segmentation models is impractical.
Instead, we leverage the Segment Anything Model (SAM)~\cite{kirillov2023segment}, a foundation model with strong zero-shot generalization, to generate high-quality object masks.
By filtering optical flow within object regions, pixel-level motion is restricted to accurate foreground areas and better reflects true target dynamics rather than background interference.
Beyond motion refinement, segmentation masks also provide alternative spatial anchors for low-confidence detections. Under partial occlusion or motion blur, bounding-box geometry and appearance embeddings often degrade, whereas segmentation masks preserve more accurate object boundaries and pixel-level spatial locations.
These observations motivate mask-based association strategies that complement traditional box-level matching.
By combining segmentation masks with per-pixel optical flow, we obtain robust pixel-level cues that improve tracking performance in complex scenarios, as illustrated in Fig.~\ref{fig: motivation}(b).

Based on these insights, we propose the SAM meets Optical Flow Tracker (SAMOFT), a novel MOT framework that leverages pixel-level cues to enhance trajectory association robustness under challenging conditions.
Specifically, SAMOFT comprises four key components:
i) Pixel Motion Matching (PMM), which refines instance-level prediction using instantaneous per-pixel motion to address Kalman filter failures under nonlinear motion;
ii) Centroid Distance Matching (CDM), designed for low-confidence detections affected by occlusion, using mask-based centroid distance for reliable association;
iii) Distribution-Based Correction (DBC), which models long-tailed motion caused by bounding-box deformation and adaptively corrects box geometry based on optical flow magnitude statistics; and
iv) Cluster-Aware ReID (CA-ReID), which improves appearance feature representation through a selective update strategy based on detection clusters formed via IoU connectivity.
While maintaining the simplicity of the tracking-by-detection framework, our method leverages the strong generalization ability of foundation models to enable robust, training-free tracking with existing tools.

We evaluate SAMOFT on DanceTrack~\cite{sun2022dancetrack}, a challenging MOT benchmark characterized by complex target motion. The competitive performance of SAMOFT validates the effectiveness of our motivation and proposed modules. Furthermore, results on the widely used MOT17~\cite{milan2016mot16} and MOT20~\cite{dendorfer2020mot20} benchmarks demonstrate the generalization ability of our method across diverse tracking scenarios.

Our main contributions are summarized as follows:
\begin{itemize}
\item We propose SAMOFT, a tracking framework that leverages pixel-level cues derived from segmentation and optical flow to enhance trajectory association robustness in multi-object tracking. 
By exploiting instantaneous foreground pixel motion and statistics-based motion correction, SAMOFT complements conventional instance-level observations and improves robustness under complex motion patterns.

\item To refine coarse instance-level motion predictions, we design the PMM module to exploit foreground pixel motions. 
The CDM module leverages mask centroids to approximate object centers while reducing geometric interference from background pixels. 
The DBC module models historical trajectory statistics to detect and correct motion prediction errors under long-tailed patterns. 
Meanwhile, we introduce CA-ReID to improve the stability and discriminative power of trajectory appearance features.

\item Experimental results on three popular MOT benchmarks validate the effectiveness of our design and the superior association robustness of SAMOFT.
\end{itemize}

\section{Related Work}
\label{sec:relatedwork}
Multi-object tracking (MOT) has received extensive attention in recent years. 
Tracking-by-detection is one of the dominant MOT paradigms, where object trajectories are constructed by associating detections across frames using various matching cues. 
Recent studies have also explored end-to-end MOT architectures that jointly model detection and association~\cite{zeng2022motr,zhang2023motrv2,10091759,10753449}. However, these approaches fall outside the scope of this paper, which focuses on improving association cues within the tracking-by-detection paradigm.
In the following, we review the most relevant studies from four perspectives: tracking-by-detection frameworks, matching cues for trajectory association, SAM-assisted MOT methods, and point-based tracking approaches.

\subsection{Tracking-by-Detection}
\label{sec:relatedwork_tbd}
The widely adopted tracking-by-detection paradigm~\cite{bewley2016simple,zhang2022bytetrack,cao2023observation} decouples MOT into object detection and trajectory association. With the advancement of robust object detectors~\cite{ren2015faster,ge2021yolox,wang2023yolov7}, current research mainly focuses on designing effective association strategies.
Typically, the association step constructs a cost matrix between detections and trajectories, followed by matching via the Hungarian algorithm~\cite{kuhn1955hungarian}.
To achieve accurate data association, many methods focus on state modeling, where motion prediction–based spatial cues and ReID-based appearance cues are utilized.
For motion modeling, the Kalman filter (KF)~\cite{kalman1960new} is the most widely used predictor, which iteratively updates its state using matched observations and predicts target motion in each frame.
Several studies improve the KF formulation~\cite{du2023strongsort,jung2024conftrack,10439025} or design neural KFs~\cite{li2024sampling,du2024ikun} to enhance prediction capability, while others adopt alternative predictors~\cite{lv2024diffmot} to replace KF.
For appearance modeling, ReID networks extract appearance embeddings from detection patches frame-by-frame~\cite{wojke2017simple}. The trajectory appearance feature is updated online using exponential moving average~\cite{wang2020towards} once a successful association is established. Studies have explored improvements through stronger feature extractors~\cite{aharon2022bot,2023ghost}, improved feature update mechanisms~\cite{maggiolino2023deep,zhang2023stat}, or affinity computation based on transformed representations~\cite{cao2025topic}.
Beyond state modeling, some methods explore adaptive weighting strategies~\cite{maggiolino2023deep} or collaboration mechanisms between motion and appearance cues~\cite{aharon2022bot} to obtain more discriminative cost matrices. Other methods focus on association strategies, such as cascade association~\cite{wojke2017simple,zhang2022bytetrack}, penalty terms~\cite{shim2025focusing}, or alternative matching algorithms~\cite{shim2025focusing}. Additionally, several works investigate post-processing steps, including matched-pair correction~\cite{huang2024deconfusetrack} and trajectory interpolation~\cite{zhang2022bytetrack,du2023strongsort}.
Our method follows the tracking-by-detection paradigm and adopts the same motion predictor, appearance feature extractor, and matching algorithm as prior methods~\cite{cao2023observation,yang2024hybrid}. Unlike previous approaches that rely solely on historical state modeling, we explicitly incorporate instantaneous target changes to compensate for motion prediction errors and suppress noise in appearance updates.

\subsection{Matching Cues in MOT}
Association cues play a fundamental role in accurate multi-object tracking.
By integrating diverse cues, trackers can maintain robust performance across varied scenarios.
Beyond the motion and appearance cues discussed in Section~\ref{sec:relatedwork_tbd}, additional instance-level cues such as detection confidence~\cite{jung2024conftrack,yang2024hybrid}, motion direction~\cite{cao2023observation}, object height~\cite{yang2024hybrid}, and pseudo-depth~\cite{dendorfer2022quo,liu2023sparsetrack} have been incorporated to enrich object modeling and improve matching reliability.
In recent years, optical flow has emerged as an effective cue for capturing additional motion information.
DK-flow-tracking~\cite{chen2019dkflow} integrates flow networks to estimate average flow of near-camera objects and corner-point motion for distant small-scale targets.
FOLT~\cite{yao2023folt} encodes optical flow maps centered on past target locations to predict inter-frame displacement.
More recently, OFTrack~\cite{song2025oftrack} uses a shared backbone for detection and optical flow, where a flow head predicts multi-frame object motion.
In the UAV tracking field, MoDe-Track~\cite{Song2026MoDe} decomposes dense optical flow into background and foreground components to compensate for camera motion and guide feature propagation.
Despite their effectiveness in capturing motion dynamics, most approaches aggregate flow information at the instance level, which may limit robustness under severe shape deformation or occlusion.
In contrast, SAMOFT combines segmentation and dense optical flow to exploit pixel-level motion cues and constructs explicit pixel-level motion affinity metrics, enabling finer-grained motion guidance and more flexible association in challenging tracking scenarios.

\subsection{SAM-Assisted MOT}
SAM has demonstrated strong effectiveness in tracking tasks when applied in a training-free manner~\cite{11351313}.
Among SAM-assisted MOT methods, MASA~\cite{li2024matching} leverages SAM’s instance-level priors to train an adapter for generalizable association, achieving zero-shot matching across open-vocabulary and driving-scene MOT benchmarks.
SAM2MOT~\cite{jiang2025sam2mot} incorporates trajectory management and an occlusion-aware correction mechanism into the SAM2 tracker, significantly improving performance but requiring sequence-specific hyperparameter tuning to handle distribution shifts.
DVA~\cite{chen2024delving} employs SAM for data augmentation by segmenting objects, perturbing backgrounds, and recomposing scenes to enhance the model’s attention to pedestrian features.
In contrast, we utilize SAM in a training-free manner to generate pixel-level motion cues rather than instance-level priors, maintaining the simplicity of the tracking-by-detection framework while improving association robustness in complex motion scenarios.

\subsection{Point Tracking in MOT}
To handle highly dynamic MOT scenarios, NetTrack~\cite{zheng2024nettrack} leverages keypoints as fine-grained cues for association and builds a point tracker on top of CoTracker~\cite{karaev2024cotracker}. 
However, NetTrack selects keypoints using a fixed uniform sampling strategy, which may not align with task-relevant pixels on the target. 
Consequently, the sampled keypoints are susceptible to sampling-induced errors, potentially limiting accurate motion modeling.
In contrast, our approach derives pixel-level motion and spatial cues from all target pixels by integrating SAM with a dense optical flow model. 
This segmentation-guided dense modeling strategy enables more complete motion perception and more adaptive keypoint sampling, thereby improving robustness under challenging dynamic conditions.

\section{Method}\label{sec:method}
\begin{figure*}[t]
\centering
\includegraphics[width=0.88\textwidth]{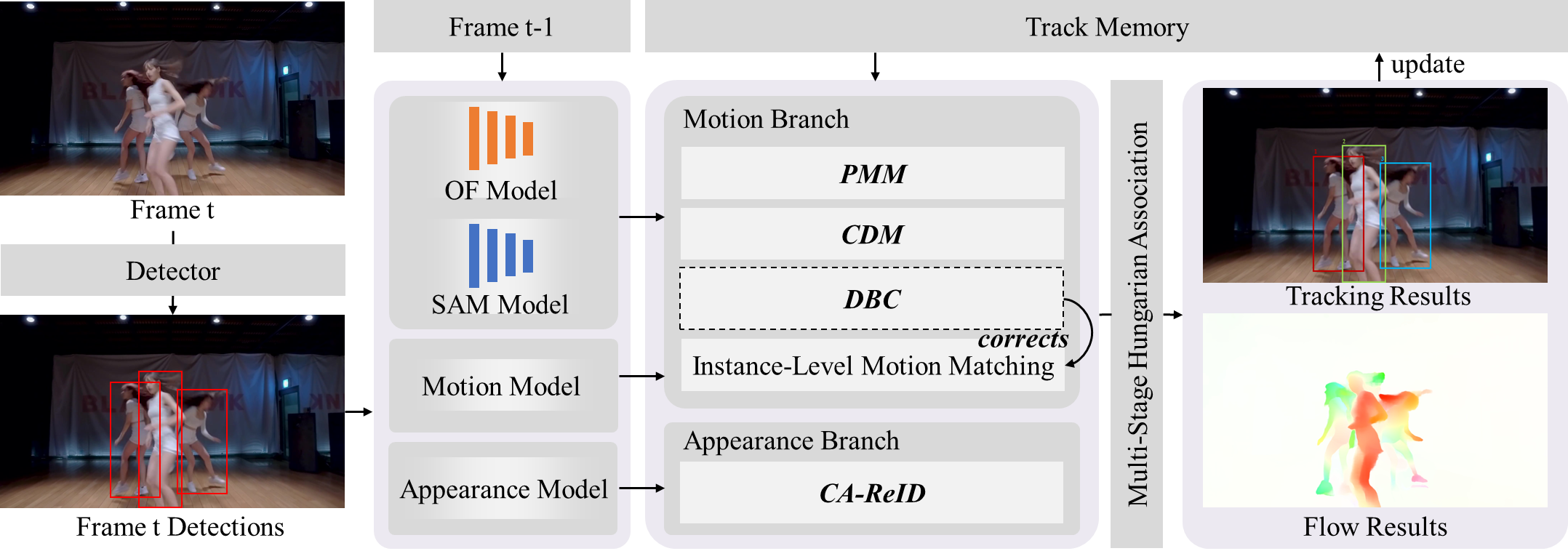}
\caption{Overall pipeline of SAMOFT. In the motion branch, PMM and CDM use SAM and an optical flow model to provide pixel-level matching cues, while DBC models long-tailed object motion and corrects trajectory motion states accordingly. In the appearance branch, CA-ReID selects the highest-quality embedding from each detection cluster to obtain discriminative appearance cues.}
\label{fig:pipeline}
\end{figure*}

\subsection{Overview of SAMOFT}

We propose SAMOFT, which leverages the Segment Anything Model (SAM)~\cite{kirillov2023segment} and an optical flow model~\cite{wang2024sea} to provide pixel-level auxiliary cues for tracking under complex motion scenarios, as illustrated in Fig.~\ref{fig:pipeline}. 
An object detector first produces instance bounding boxes in the current frame for trajectory association. 
Unlike conventional approaches that rely solely on the Kalman Filter (KF) motion model and appearance embeddings to represent each instance, we incorporate an optical flow model to estimate pixel-level motion. 
Meanwhile, SAM is used to ensure that the selected pixels belong to the target object and to provide additional association cues for potentially occluded targets.

The SAM- and optical-flow–driven modules include Pixel Motion Matching (PMM) and Centroid Distance Matching (CDM). 
SAMOFT also introduces a Distribution-Based Correction (DBC) module that models and corrects long-tailed motion events using pixel motion statistics. 
In addition, a Cluster-Aware ReID (CA-ReID) module provides more discriminative appearance cues. 
Finally, the costs from CA-ReID, PMM, CDM, and the DBC-corrected instance-level motion cues are fused to construct the final cost matrix for Hungarian association. 
Algorithm~\ref{alg:samoft} summarizes the overall pipeline of SAMOFT.

\begin{algorithm}[t]
\caption{Overall Pipeline of SAMOFT}
\label{alg:samoft}
\KwIn{Current trajectories $\mathcal{T}_{t-1}$; current detections $\mathcal{D}_{t}$}
\KwOut{Updated trajectories $\mathcal{T}_{t}$}

\For{each track $T_i \in \mathcal{T}_{t-1}$}{
    Estimate KF prediction $X_i$, mask $\mathbf{M}_i$, and pixel locations $\mathbf{P}_i^{t}$\;
    Load appearance feature $\mathbf{F}_i$\;
}

\For{each detection $D_j \in \mathcal{D}_t$}{
    Extract ReID embedding $\mathbf{F}_j$ and mask $\mathbf{M}_j$\;
}

\For{each track $T_i \in \mathcal{T}_{t-1}$ and detection $D_j \in \mathcal{D}_t$}{
    Compute $C_{ILM}(T_i, X_i, D_j)$ using IoU and OCM\;
    $C_{ILM}(T_i, D_j) \leftarrow DBC(C_{ILM}(T_i, D_j))$\;
    Compute $C_{P}(\mathbf{P}_i^{t}, \mathbf{M}_j)$\;
    Compute $C_{C}(\mathbf{M}_i, \mathbf{M}_j)$\;
    Compute appearance cost $C_{CA-ReID}(\mathbf{F}_i, \mathbf{F}_j)$\;
}

Perform multi-stage Hungarian matching with the above costs to obtain $\mathrm{Matches}_t$\;

\For{each matched pair $(T_i, D_j) \in \mathrm{Matches}_t$}{
    Update KF and DBC statistics\;
}

Build IoU graph for matched detections\;
Obtain clusters $\mathrm{Clusters}_t$ via connected components\;

\For{each cluster $c_i \in \mathrm{Clusters}_t$}{
    Select highest-confidence detection $D_{\text{top}} \in c_i$\;
    Update trajectory appearance using $\mathbf{F}_{\text{top}}$\;
}

Update $\mathcal{T}_{t}$ with new-born trajectories and matched pairs\;

\Return{$\mathcal{T}_{t}$}\;
\end{algorithm}

\subsection{Pixel Motion Matching}

\begin{figure}[t]
\centering
\includegraphics[width=1.\linewidth]{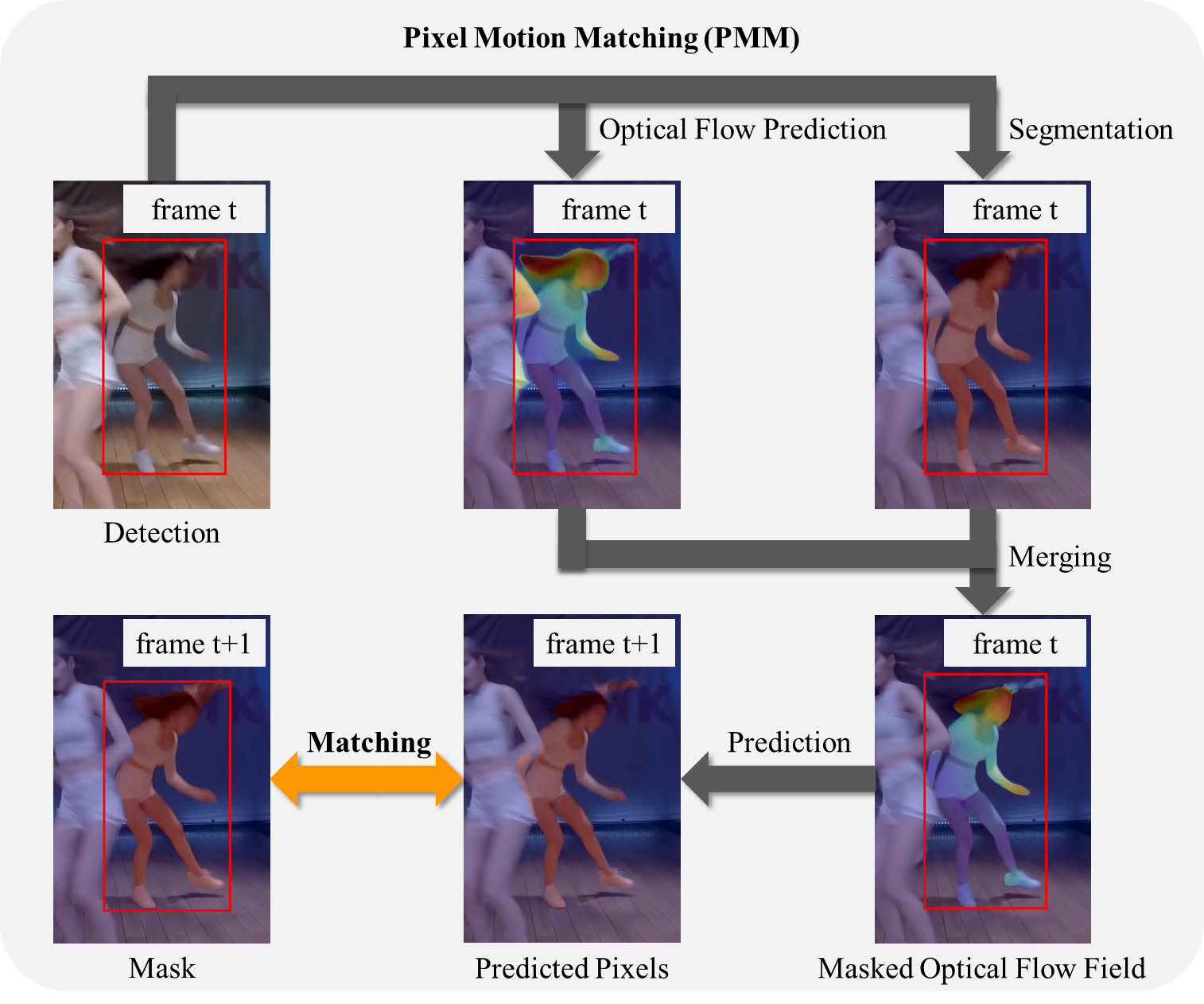}
\caption{Illustration of Pixel Motion Matching (PMM). Target pixel positions in the next frame are predicted using segmentation masks and optical flow, and the predicted pixels are matched with detection masks.}
\label{fig:ILLUS_PMM}
\end{figure}

Most tracking-by-detection methods rely on instance-level motion predictions (i.e., track bounding boxes) from the KF to match detections. However, bounding-box prediction is sensitive to target deformation. In non-rigid scenarios (e.g., human tracking), body movements cause significant shape variations that degrade prediction robustness. Moreover, the KF state transition and observation models assume linear motion, which leads to inaccuracies under nonlinear dynamics, and these errors may accumulate over time.

To address these issues, we propose Pixel Motion Matching (PMM), which leverages SAM to extract target pixels within track and detection boxes and uses an optical flow model to estimate instantaneous pixel motion between consecutive frames, as illustrated in Fig.~\ref{fig:ILLUS_PMM}. By extracting pixels using a reliable segmentation model (SAM in our implementation), PMM reduces the effect of bounding-box deformation. In addition, pixel-level motion estimated from optical flow is independent of historical track states, providing a finer and deformation-invariant motion representation.

For all trajectories and detections, we first apply SAM to obtain the target pixel masks within their bounding boxes:
\begin{equation}
\label{eq:sam_mask}
\mathbf{M}_i = \text{SAM}(\mathbf{T}_i), \quad 
\mathbf{M}_j = \text{SAM}(\mathbf{D}_j),
\end{equation}
where $\mathbf{T}_i$ and $\mathbf{D}_j$ denote the bounding boxes of trajectory $i$'s last observation and detection $j$, and $\text{SAM}(\cdot)$ extracts the corresponding target pixel mask.

Meanwhile, the optical flow model computes the flow field between consecutive frames. By adding the flow displacement to trajectory pixels, we estimate their locations in the next frame:
\begin{equation}
\label{eq:flow_predict}
\mathbf{P}_i^{t+1} = \mathbf{P}_i^{t} + \text{Flow}(\mathbf{P}_i^{t}),
\end{equation}
where $\mathbf{P}_i^t$ denotes the pixel coordinates of $\mathbf{M}_i$ at frame $t$, and $\text{Flow}(\mathbf{P}_i^{t})$ represents the optical flow displacement between frames $t$ and $t+1$.

Finally, we construct the PMM association cost matrix as
\begin{equation}
\label{eq:pmm_cost}
\mathbf{C}_{P}(i,j) = -
\frac{|\mathbf{P}_i^{t+1} \cap \mathbf{M}_j|}{|\mathbf{P}_i^{t+1}|},
\end{equation}
where $\mathbf{C}_{P}(i,j)$ denotes the PMM cost between trajectory $i$ and detection $j$, computed as the negative proportion of predicted trajectory pixels falling within the detection mask.

\subsection{Centroid Distance Matching}

\begin{figure}[t]
\centering
\includegraphics[width=0.72\linewidth]{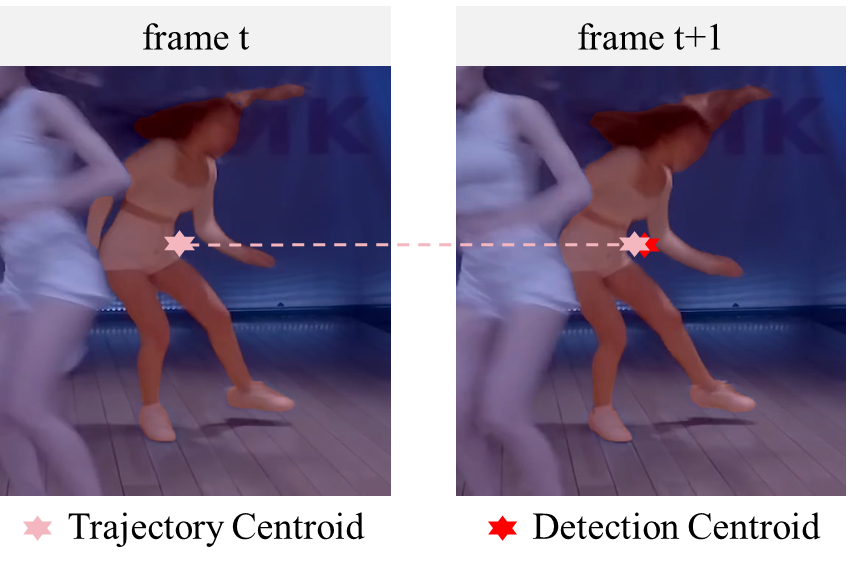}
\caption{Illustration of mask centroids of a trajectory in the previous frame and a detection in the current frame. The centroids may lie outside the mask pixels.}
\label{fig:ILLUS_CDM}
\end{figure}

Previous work~\cite{zhang2022bytetrack} shows that low-confidence detections may still correspond to valid targets, although low confidence often indicates occlusion or motion blur. 
To better utilize such detections, we propose Centroid Distance Matching (CDM), which uses target masks to assist association. 
Unlike conventional distance matching based on bounding-box centers, CDM computes centroids from visible mask pixels, enabling more robust localization under partial occlusion. 
This design provides greater flexibility when handling complex motion patterns.

For trajectory $i$ and detection $j$, we compute the centroids of their masks, as illustrated in Fig.~\ref{fig:ILLUS_CDM}, and measure their Euclidean distance:
\begin{equation}
\label{eq:centroid_dist}
d_{ij} = \left\| \text{Centroid}(\mathbf{M}_i) - \text{Centroid}(\mathbf{M}_j) \right\|_2.
\end{equation}

To further improve discriminability, we normalize the distance matrix using a Gaussian-based transformation, yielding the CDM association cost:
\begin{equation}
\label{eq:cdm_cost}
\mathbf{C}_{C}(i,j) = -e^{1 - \frac{d_{ij}}{\text{Min}(d_i)}},
\end{equation}
where $\text{Min}(d_i)$ denotes the smallest centroid distance between trajectory $i$ and its nearest detection. 
This normalization prioritizes nearby detections while enlarging distance differences among overlapping detections and leaving distant detections largely unaffected.

\subsection{Distribution-Based Correction}

\begin{figure}[t]
\centering
\includegraphics[width=1.\linewidth]{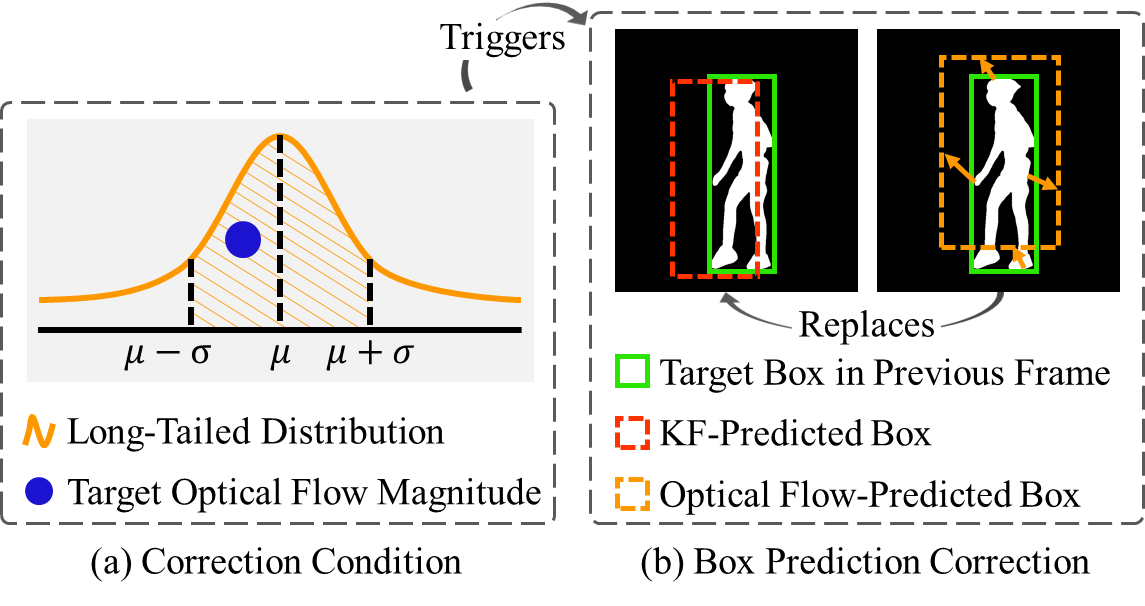}
\caption{Illustration of Distribution-Based Correction (DBC). Based on the historical distribution and current value of optical flow magnitude, IoU correction is triggered adaptively.}
\label{fig:ILLUS_DBC}
\end{figure}

We observe that when a target undergoes long-tailed motion, its optical flow magnitude often deviates from its normal periodic range. 
During significant bounding-box deformation, strong limb movements may produce peaks in flow magnitude. 
Based on this observation, we propose a training-free Distribution-Based Correction (DBC) module. 
DBC statistically models the optical flow magnitude distribution of target pixels to capture long-tailed motion patterns. 
During association, it adaptively corrects KF-predicted bounding boxes using pixel-level motion cues.

The optical flow magnitude of pixel $p$ is defined as
\begin{equation}
\label{eq:magnitude}
m(p) = \sqrt{u(p)^2 + v(p)^2},
\end{equation}
where $u(\cdot)$ and $v(\cdot)$ denote the horizontal and vertical flow components. 

After a successful association, we compare the IoU between the optical-flow–predicted bounding box and the KF prediction. 
If the flow-based IoU exceeds the KF-based IoU by a threshold $\tau_{D}$, we update the long-tailed flow distribution statistics:
\begin{equation}
n_{new} = n_{prev} + n_{t},
\label{eq:flow_stats_update_n}
\end{equation}
\begin{equation}
\mu_{new} = \frac{n_{prev}\mu_{prev} + n_{t}\mu_{t}}{n_{prev}+n_{t}},
\label{eq:flow_stats_update_mu}
\end{equation}
\begin{equation}
\sigma_{new} =
\frac{\sqrt{n_{prev}\sigma_{prev}^2 + n_{t}\sigma_{t}^2 +
\frac{n_{prev}n_{t}(\mu_{prev}-\mu_{t})^2}{n_{prev}+n_{t}}}}
{n_{prev}+n_{t}},
\label{eq:flow_stats_update_sigma}
\end{equation}
where $(n,\mu,\sigma)$ denote the number of pixels, mean, and standard deviation of the flow magnitude distribution after update, before update, and from the current mask, respectively.

As illustrated in Fig.~\ref{fig:ILLUS_DBC}(a), during association, if the mean magnitude of a trajectory mask falls within one standard deviation of the distribution mean, DBC correction is triggered. 
The flow-based predicted bounding box is obtained from the minimum and maximum pixel coordinates of the predicted trajectory pixels $p_i \in P_i$, as shown in Fig.~\ref{fig:ILLUS_DBC}(b). 
The resulting box replaces the KF prediction for IoU computation if it produces a higher score. 
Since optical flow estimation is instantaneous and independent of historical states, it effectively compensates for KF prediction errors.

\subsection{Cluster-Aware ReID}

The effectiveness of appearance cues for association has been widely demonstrated~\cite{aharon2022bot,maggiolino2023deep,yang2024hybrid}. 
To further improve robustness, we incorporate an independent appearance branch, as shown in Fig.~\ref{fig:pipeline}. 
Detection patches are fed into an appearance network to extract ReID features, and trajectory features are updated online using an exponential moving average. 
The appearance cost is then computed using cosine distance between trajectory and detection embeddings.

Different from previous approaches, our Cluster-Aware ReID (CA-ReID) introduces a clustering step before feature update. 
Detections in the current frame are grouped based on IoU connectivity. 
Specifically, we construct an undirected graph where each node represents a detection, and edges connect detections whose IoU exceeds a threshold (0.3 in our experiments). 
Connected components in this graph are treated as detection clusters.

For each cluster, only the embedding of the highest-confidence detection is used to update the trajectory feature. 
This strategy reduces noise in feature updates and improves discriminative capability.

\subsection{Multi-Stage Hungarian Association}

Our multi-stage Hungarian association follows the baseline framework, with an additional BYTE~\cite{zhang2022bytetrack} association stage when CDM is enabled. 
Specifically, the association module consists of three stages: 
(i) high-confidence detection matching, 
(ii) low-confidence association via BYTE, and 
(iii) trajectory recovery via OCR (as in OC-SORT). 

The instance-level motion cost inherited from OC-SORT~\cite{cao2023observation} combines spatial localization and velocity direction consistency:
\begin{equation}
\mathbf{C}_{ILM} = \mathbf{C}_{IoU} + \lambda_{OCM} \mathbf{C}_{OCM}.
\label{eq:ocsort_cost}
\end{equation}
SAMOFT retains this formulation while refining $C_{IoU}$ through DBC and incorporating additional pixel-level and appearance cues:
\begin{equation}
\mathbf{C}_{final} =
\lambda_{ILM} \cdot DBC(\mathbf{C}_{ILM}) + \mathbf{C}_{ours},
\label{eq:samoft_cost}
\end{equation}
\begin{equation}
\mathbf{C}_{ours} =
\mathbf{C}_{motion} + \lambda_{appr} \cdot \mathbf{C}_{CA-ReID},
\label{eq:samoft_cost_new}
\end{equation}
\begin{equation}
\mathbf{C}_{motion} =
\lambda_{P} \mathbf{C}_{P} + \lambda_{C} \mathbf{C}_{C},
\label{eq:samoft_motion}
\end{equation}
where $\lambda_{P}$ and $\lambda_{C}$ are empirically determined weights, while $\lambda_{OCM}$, $\lambda_{ILM}$, and $\lambda_{appr}$ follow the settings in prior work~\cite{cao2023observation,yang2024hybrid}.

\section{Experiments}
\label{sec: Experiments}
\label{sec:experiments}

\subsection{Experiment Setups}
\subsubsection{Datasets and Metrics}
We conduct experiments on several public datasets to evaluate the generalization capability of our method, including DanceTrack, MOT17, and MOT20.
DanceTrack consists of 100 video sequences, with 40 for training, 25 for validation, and 35 for testing.
MOT17 contains 14 sequences for pedestrian tracking.
MOT20 includes 4 training and 4 testing sequences for pedestrian tracking and features denser scenes with higher target density than MOT17.
We emphasize results on DanceTrack due to its diverse motion patterns and long target durations.
For quantitative evaluation, we adopt Higher Order Tracking Accuracy (HOTA)~\cite{luiten2021hota} as the primary metric. Meanwhile, IDF1~\cite{ristani2016performance}, AssA~\cite{luiten2021hota}, and AssR~\cite{luiten2021hota} assess association quality, while MOTA~\cite{bernardin2008evaluating} and DetA~\cite{luiten2021hota} reflect detection-related performance.

\subsubsection{Implementation Details}
\begin{table*}[t]
        \caption{Performance comparison on the DanceTrack test set under the private detection protocol. Methods in the bottom block use the same detections. The gray block indicates our integration. The top three results in each block are highlighted in \textcolor{red}{red}, \textcolor{blue}{blue}, and \textcolor{green}{green}.}
        \centering
                
         \begin{tabular}{l | c | c c c c c}
        \toprule
             
             Tracker & Venue & HOTA↑ & IDF1↑ & MOTA↑ & DetA↑ & AssA↑\\
             \midrule
             \multicolumn{7}{l}{\textit{Different Detection Settings:}}\\
              GTR\cite{zhou2022global}& CVPR22                        & 48.0 & 50.3 & 84.7 & 72.5 & 31.9 \\
              MOTR\cite{zeng2022motr}&  ECCV22                         & 54.2 & 51.5 & 79.7 & 73.5 & 40.2 \\
              SUSHI\cite{cetintas2023unifying}& CVPR23                & \textcolor{green}{63.3} & \textcolor{green}{63.4} & 88.7 & 80.1 & \textcolor{green}{50.1} \\
              MOTRv2\cite{zhang2023motrv2}& CVPR23                    & \textcolor{red}{69.9} & \textcolor{blue}{71.7} & \textcolor{red}{91.9} & \textcolor{red}{83.0} & \textcolor{blue}{59.0} \\   
              DiffusionTrack\cite{luo2024diffusiontrack}    & AAAI24 & 52.4 & 47.5 & \textcolor{green}{89.3} & \textcolor{blue}{82.2} & 33.5  \\
              MOTIP\cite{gao2025multiple} &  CVPR25                    & \textcolor{blue}{69.6} & \textcolor{red}{74.7} & \textcolor{blue}{90.6} & \textcolor{green}{80.4} & \textcolor{red}{60.4} \\
             \midrule
             \multicolumn{7}{l}{\textit{Same Detection:}}\\           
             ByteTrack\cite{zhang2022bytetrack}&ECCV22     & 47.3 & 52.5 & 89.5 & 71.6 & 31.4\\
             FineTrack\cite{ren2023focus}& CVPR23          & 52.7 & 59.8 & 89.9 & 72.4 & 38.5\\             
             GHOST\cite{2023ghost}         &CVPR23         & 56.7 & 57.7 & 91.3 & 81.1 & 39.8\\
             StrongSORT++\cite{du2023strongsort}& TMM23    & 55.6 & 55.2 & 91.1 & 80.7 & 38.6 \\             
             STAT\cite{zhang2023stat}& TMM23               & 60.4 & 61.4 & 91.6 & 81.1 & 45.1 \\             
             Deep OC-SORT\cite{maggiolino2023deep}&ICIP23  & 61.3 & 61.5 & \textcolor{green}{92.3} & \textcolor{green}{82.2} & 45.8  \\
                 AHOR-ReID\cite{jin2024ahor} & TCSVT24         & 60.7 & 59.9 & 91.5 & \textcolor{blue}{82.3} & 44.9  \\           
             CMTrack\cite{Shim2024CMTrack} & ICIP24        & 61.8 & 63.3 & \textcolor{green}{92.3} & \textcolor{green}{82.2} & 45.8  \\
             DiffMOT\cite{lv2024diffmot} & CVPR24          & 63.4  & 64.0 & \textcolor{blue}{92.7} & \textcolor{red}{82.5} & 48.8  \\
             GeneralTrack\cite{qin2024towards}& CVPR24     & 59.2 & 59.7 & 91.8 & 82.0 & 42.8 \\
             UCMCTrack+\cite{yi2024ucmctrack} & AAAI24     & 63.6 & 65.0 & 88.9 & - &  51.3\\              
             CoNo-Link\cite{gao2024multi}    &  AAAI24     & 63.8 & 64.1 & 89.7 & 80.2 & 50.7 \\
             Hybrid-SORT-ReID\cite{yang2024hybrid}& AAAI24 & \textcolor{green}{65.7} & \textcolor{green}{67.4} & 91.8 & - & - \\              
             DfTrack-Hybrid\cite{11007673}& TCSVT25        & 65.5 & \textcolor{blue}{68.2} & \textcolor{blue}{92.7} & 82.1 & \textcolor{green}{52.4} \\
             TOPICTrack\cite{cao2025topic}& TIP25          & 58.3 & 58.4 & 90.9 & 80.7 & 42.3 \\
             OFTrack-ReID\cite{song2025oftrack}& AAAI25    & 63.4 & 65.6 & 91.2 & 82.1 & 48.7 \\
             TrackTrack\cite{shim2025focusing}& CVPR25     & \textcolor{blue}{66.5} &  67.8 & \textcolor{red}{93.6} & - & \textcolor{blue}{52.9} \\
             
             \rowcolor{gray!20} OC-SORT\cite{cao2023observation}& CVPR23       & 54.6 & 54.6 & 89.6 & 80.4 & 40.2 \\
             \rowcolor{gray!20} \textbf{OC-SORT+SAMOFT}& Ours                             & 63.8 & 65.9 & 91.6 & 82.0  & 49.8 \\      
            \rowcolor{gray!20} Hybrid-SORT\cite{yang2024hybrid} & AAAI24     & 62.2 & 63.0 & 91.6 & - & - \\
            \rowcolor{gray!20} \textbf{Hybrid-SORT+SAMOFT}& Ours                             & \textcolor{red}{68.2} & \textcolor{red}{71.5} &  91.5 & 81.7 & \textcolor{red}{57.1} \\            
            \bottomrule
        \end{tabular}
        \label{tab:benchmark}        
\end{table*}

\begin{table}[t]
        \caption{Performance comparison on the MOT17 test set under the private detection protocol. Methods in the bottom block use the same detections. Gray rows indicate our integration. Top-3 results in each block are highlighted in \textcolor{red}{red}, \textcolor{blue}{blue}, and \textcolor{green}{green}.}
        \centering
                
         \begin{tabular}{l | c | c c c }
        \toprule             
             Tracker & Venue & HOTA↑ & IDF1↑ & MOTA↑ \\
             \midrule
             \multicolumn{5}{l}{\textit{Different Detection Settings:}}\\
              GTR\cite{zhou2022global}& CVPR22                        & 59.1 & 71.5 & 75.3  \\
              MOTR\cite{zeng2022motr}&  ECCV22                        & 57.8 & 68.6 & 73.4  \\
              SUSHI\cite{cetintas2023unifying}& CVPR23            & \textcolor{red}{66.5} & \textcolor{red}{83.1} & \textcolor{blue}{81.1}  \\
              MOTRv2\cite{zhang2023motrv2} & CVPR23              & 62.0 & 75.0 & 78.6 \\
              SMILEtrack\cite{wang2024smiletrack} & AAAI24       & \textcolor{blue}{65.3} & \textcolor{blue}{80.5} & \textcolor{blue}{81.1} \\     
              DiffusionTrack\cite{luo2024diffusiontrack} & AAAI24 & 60.8 & 73.8 & 77.9 \\             
              TGFormer\cite{zeng2025tgformer}& AAAI25            & 60.3 & 72.0 & 74.9  \\
              FDTracker\cite{fu2025foundation}& AAAI25           & \textcolor{green}{64.2} & \textcolor{green}{79.2} & \textcolor{red}{81.8}  \\
              LA-MOTR\cite{wang2025lamotr}& ICCV25                &  62.6 & 73.8 & \textcolor{green}{80.7}  \\
              MOTIP\cite{gao2025multiple}&  CVPR25                    & 59.3 & 71.3 & 75.3 \\
                          
             \midrule
             \multicolumn{5}{l}{\textit{Same Detection:}}\\  
             ByteTrack\cite{zhang2022bytetrack}& ECCV22         & 63.1 & 77.3 & 80.3 \\
             GHOST\cite{2023ghost}& CVPR23                      & 62.8 & 77.1 & 78.7 \\
             MotionTrack\cite{qin2023motiontrack}  & CVPR23     & \textcolor{red}{65.1} & \textcolor{blue}{80.1} & \textcolor{red}{81.1} \\
             STAT\cite{zhang2023stat}& TMM23                    & 63.7 & 79.0 & 78.7 \\
             Hybrid-SORT-ReID\cite{yang2024hybrid}& AAAI24      & 64.0 & 78.7 & 79.9 \\
             GeneralTrack\cite{qin2024towards}& CVPR24          & 64.0 & 78.3 & \textcolor{blue}{80.6} \\
             DeconfuseTrack\cite{huang2024deconfusetrack}   &  CVPR24 & \textcolor{blue}{64.9} & \textcolor{red}{80.6} & \textcolor{green}{80.4} \\
             TOPICTrack\cite{cao2025topic}& TIP25               & 63.9 & 78.7 & 78.8 \\    
             OFTrack-ReID\cite{song2025oftrack}& AAAI25         & 64.1 & 78.8 & 80.1 \\
             
             \rowcolor{gray!20}OC-SORT\cite{cao2023observation}& CVPR23           & 63.2 & 77.5 & 78.0 \\
             \rowcolor{gray!20}\textbf{OC-SORT+SAMOFT}& Ours                              & 64.0 & 78.8 & 79.9 \\       
             \rowcolor{gray!20}Hybrid-SORT\cite{yang2024hybrid} & AAAI24             & 63.6 & 78.4 & 79.3 \\
             \rowcolor{gray!20}\textbf{Hybrid-SORT+SAMOFT}& Ours                             & \textcolor{green}{64.3} & \textcolor{green}{79.4} & 80.0  \\            
            \bottomrule
        \end{tabular}         
        \label{tab:benchmark_mot17}        
\end{table}

\begin{table}[t]
        \caption{Performance comparison on the MOT20 test set under the private detection protocol. Methods in the bottom block use the same detections. Gray rows indicate our integration. Top-3 results in each block are highlighted in \textcolor{red}{red}, \textcolor{blue}{blue}, and \textcolor{green}{green}.}
        \centering
                
         \begin{tabular}{l | c | c c c }
        \toprule
             
             Tracker & Venue & HOTA↑ & IDF1↑ & MOTA↑ \\
             \midrule
             \multicolumn{5}{l}{\textit{Different Detection Settings:}}\\
            SUSHI\cite{cetintas2023unifying}& CVPR23              & \textcolor{red}{64.3} & \textcolor{red}{79.8} & 74.3  \\
            MOTRv2\cite{zhang2023motrv2}& CVPR23                  & \textcolor{green}{61.0} & 73.1 & \textcolor{green}{76.2} \\
            SMILEtrack\cite{wang2024smiletrack}& AAAI24          & \textcolor{blue}{63.4} & \textcolor{blue}{77.5} & \textcolor{blue}{78.2} \\  
            DiffusionTrack\cite{luo2024diffusiontrack} & AAAI24  & 55.3 & 66.3 & 72.8 \\
            FDTracker\cite{fu2025foundation}& AAAI25             & \textcolor{blue}{63.4} & \textcolor{green}{76.9} & \textcolor{red}{78.3}  \\
            TGFormer\cite{zeng2025tgformer}& AAAI25      & 54.2 & 67.1  & 70.3  \\
             
             \midrule
             \multicolumn{5}{l}{\textit{Same Detection:}}\\  
             ByteTrack\cite{zhang2022bytetrack}&ECCV22            & 61.3 & 75.2 & \textcolor{green}{77.8} \\
             GHOST\cite{2023ghost}&CVPR23                         & 61.2 & 75.2 & 73.7 \\
             MotionTrack\cite{qin2023motiontrack} & CVPR23     & 62.8  & 76.5 & \textcolor{blue}{78.0} \\
             STAT\cite{zhang2023stat}& TMM23                      & 62.5 & 76.4 & 75.5 \\
             StrongSORT++\cite{du2023strongsort}& TMM23           & 62.6 & 77.0 & 73.8 \\
             Hybrid-SORT-ReID\cite{yang2024hybrid}& AAAI24        & \textcolor{green}{63.9} & \textcolor{blue}{78.4} & 76.7 \\
             GeneralTrack\cite{qin2024towards}& CVPR24            & 61.4 & 74.0 & 77.2 \\
             DeconfuseTrack\cite{huang2024deconfusetrack}  &  CVPR24 & 63.3 & \textcolor{green}{77.6} & \textcolor{red}{78.1} \\    
             TOPICTrack\cite{cao2025topic}& TIP25                 & 62.6 & \textcolor{green}{77.6} & 72.4  \\  
             OFTrack-ReID\cite{song2025oftrack}& AAAI25           & 63.4 & 76.9 & 75.6 \\
             
             \rowcolor{gray!20}OC-SORT\cite{cao2023observation}& CVPR23            & 62.1 & 75.9 & 75.5 \\
             \rowcolor{gray!20}\textbf{OC-SORT+SAMOFT}& Ours                              & \textcolor{blue}{64.0} & \textcolor{red}{78.9} & 75.5 \\       
             \rowcolor{gray!20}Hybrid-SORT\cite{yang2024hybrid} & AAAI24   & 62.5 & 76.2 & 76.4  \\
             \rowcolor{gray!20}\textbf{Hybrid-SORT+SAMOFT} & Ours                  & \textcolor{red}{64.2} & \textcolor{red}{78.9} & 76.8  \\            
            \bottomrule
        \end{tabular}         
        \label{tab:benchmark_mot20}        
\end{table}

Unless otherwise specified, all results are reported under the \emph{private detection} protocol.
We choose OC-SORT~\cite{cao2023observation} as the primary baseline due to its simple yet extensible architecture and follow its original implementation and hyperparameter settings for instance-level motion matching.
In SAMOFT, the motion cost matrix consists of three components: the instance-level motion matching cost, the PMM cost, and the CDM cost.
Their weights are set to $\lambda_{ILM}=1.0$, $\lambda_{P}=0.2$, and $\lambda_{C}=1.0$, where $\lambda_{C}$ is applied only during the BYTE association stage.
The DBC update threshold is $\tau_{D}=0.1$.

To ensure fair comparisons, we adopt the same publicly available detection (YOLOX~\cite{ge2021yolox}) and ReID (BoT~\cite{yang2024hybrid}) models and weights, together with the same appearance weighting.
For our proposed modules, we use SEA-RAFT~\cite{wang2024sea} with weights trained on Spring~\cite{mehl2023spring} for optical flow estimation and the SAM model~\cite{kirillov2023segment} with provided weights for mask generation.
All experiments are conducted on an Intel i5-13600K CPU and an NVIDIA GeForce RTX 4090 GPU.

\subsection{Benchmark Evaluation}

We compare SAMOFT with recent state-of-the-art methods on DanceTrack, MOT17, and MOT20.
Besides the OC-SORT implementation, we also integrate our modules into Hybrid-SORT~\cite{yang2024hybrid} to evaluate their generalization across different trackers and demonstrate improvements with a stronger baseline.

\subsubsection{DanceTrack}

We compare SAMOFT with representative trackers on the DanceTrack test set to evaluate its effectiveness under complex motion scenarios.
As shown in Table~\ref{tab:benchmark}, our method achieves notable gains when built upon both OC-SORT and HybridSORT.
Specifically, integrating SAMOFT with OC-SORT improves performance by +9.2 HOTA, +11.3 IDF1, and +2.0 MOTA, indicating enhanced association accuracy and overall tracking performance.
When applied to HybridSORT, SAMOFT still yields consistent gains of +6.0 HOTA and +5.2 IDF1, demonstrating the robustness and generality of our modules across different trackers.

\begin{figure}[t]
\centering
\includegraphics[width=1.\linewidth]{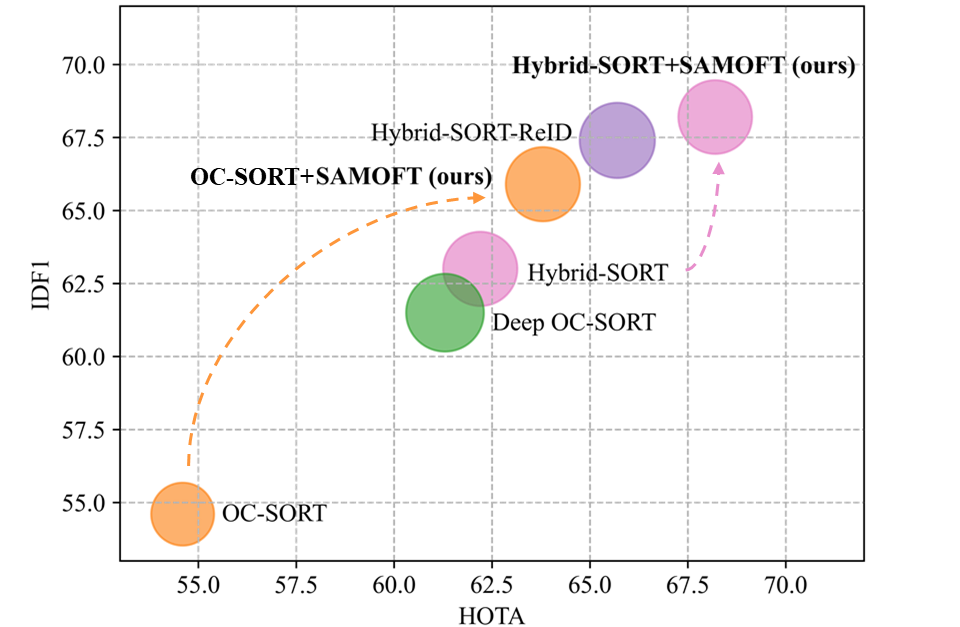}
\caption{Comparison of HOTA, IDF1, and MOTA scores on the DanceTrack test set for OC-SORT and its variants. The radius of each circle indicates the MOTA score,   the overall tracking performance of different methods.}
\label{fig:bubble_dance}
\end{figure}

These results show that our design effectively complements trackers relying mainly on instance-level motion cues by incorporating pixel-level cues, distribution-based correction, and noise-reduced appearance modeling.
As illustrated in Fig.~\ref{fig:bubble_dance}, SAMOFT surpasses the baselines and ranks among the top OC-SORT variants.
We also note that Transformer-based methods (e.g., MOTRv2 and MOTIP) achieve higher performance on DanceTrack but require expensive training.
In contrast, SAMOFT improves association robustness within the tracking-by-detection framework in a training-free manner.

\subsubsection{MOTChallenge}

We further evaluate SAMOFT on the MOT17 and MOT20 benchmarks for pedestrian tracking, as reported in Tables~\ref{tab:benchmark_mot17} and \ref{tab:benchmark_mot20}. 
SAMOFT consistently improves the corresponding baselines while achieving performance competitive with recent state-of-the-art approaches.

Specifically, when built upon OC-SORT, SAMOFT improves performance by +0.8 HOTA and +1.3 IDF1 on MOT17, and +1.9 HOTA and +3.0 IDF1 on MOT20. 
When integrated into Hybrid-SORT, SAMOFT yields gains of +0.7 HOTA, +1.0 IDF1, and +0.7 MOTA on MOT17, and +1.7 HOTA, +2.7 IDF1, and +0.4 MOTA on MOT20.

On MOT17, MotionTrack~\cite{qin2023motiontrack} and DeconfuseTrack~\cite{huang2024deconfusetrack} achieve higher HOTA scores than ours under the same detection setting. MotionTrack benefits from its ability to handle long-term occlusions, while DeconfuseTrack emphasizes refined association through post-matching correction. 
In addition, SUSHI achieves strong results on both MOT17 and MOT20 due to its graph neural network–based matcher and offline tracking pipeline that jointly optimizes association across frames. 

In contrast, SAMOFT focuses on improving motion modeling within the tracking-by-detection framework through segmentation-guided pixel-level cues and distribution-aware correction, without introducing additional training stages or complex optimization schemes. 
Although the improvements on MOT17 and MOT20 are smaller than those on DanceTrack, this observation aligns with prior analyses~\cite{yang2024hybrid,sun2022dancetrack,shuai2022large}, which show that the limited sequence diversity and predominantly linear motion patterns in these datasets reduce the benefits of advanced motion modules. 
Nevertheless, the consistent gains demonstrate the effectiveness of leveraging pixel-level cues and confirm the generalization ability of SAMOFT across both simple motion scenarios and densely crowded scenes.

\subsection{Ablation Studies}

We systematically analyze the effectiveness and design choices of SAMOFT.
First, we evaluate the contribution of each component. 
Next, we study the effect of key hyperparameters in PMM, CDM, and DBC. 
Finally, we validate several design decisions, including the selective correction strategy in DBC, the cluster-aware update strategy in CA-ReID, the choice of optical flow estimator, and the segmentation-based pixel filtering strategy.

\subsubsection{Component-Wise Analysis}

Table~\ref{tab:ablation_all} summarizes the impact of integrating the proposed components.
Each module independently improves HOTA, and their combination progressively enhances overall tracking performance.

The PMM module improves localization under nonlinear motion and partially overlapping trajectories by introducing pixel-level motion cues.
The CDM module increases the utilization of low-confidence detections via mask-centroid matching while maintaining reliable associations.
The DBC module models the distribution of optical flow magnitudes to statistically correct bounding box geometry when abnormal motion occurs, strengthening motion robustness.

Building upon these motion-level refinements, the CA-ReID module further improves ID consistency by reducing noise accumulation in trajectory-level appearance features.
With all components combined, SAMOFT achieves substantial improvements over the baseline, with gains of +8.4 HOTA, +9.2 IDF1, and +1.9 MOTA, demonstrating strong association robustness in complex motion scenarios.
We also verify the generalization of our design by integrating the modules into Hybrid-SORT, where consistent improvements are observed without additional tuning.

\begin{table}[t]
\caption{Component ablation study on the DanceTrack validation set. Each proposed module progressively improves tracking and association performance.}           
        \centering
        
        \begin{tabular}{c c c c | c c c}        
        \toprule
        PMM & CDM & DBC & CA-ReID & HOTA↑ & IDF1↑  & MOTA↑\\
        \midrule
        \multicolumn{4}{l|}{\textit{OC-SORT Implementation:}}   &  &  &    \\  
          & &        &                                                    & 52.2 & 51.9 & 87.3  \\
        \Checkmark &  & &                                                & 53.3 & 53.7 & 87.4  \\
        &  \Checkmark  &   &                                  & 53.5 & 53.3 & 88.8  \\
         & &\Checkmark  &                                                & 52.6 & 52.5 & 87.5  \\
        \Checkmark &    & \Checkmark &                         & 53.6 & 54.2 & 87.5  \\
        \Checkmark &  \Checkmark  &  &                         &  54.8  & 55.4 & 88.7  \\
        \Checkmark &\Checkmark  & \Checkmark&                 & 55.8 &  56.2 & 88.9  \\
         \rowcolor{gray!20}\Checkmark& \Checkmark  & \Checkmark & \Checkmark      & \textbf{60.6} & \textbf{61.1} & \textbf{89.2} \\
       \midrule
        \multicolumn{4}{l|}{\textit{Hybrid-SORT Implementation:}}   &  &  &    \\       
         &  &  & &59.2  & 60.4 & 89.5 \\
        &        &          & \Checkmark & 62.9 & 64.6 &\textbf{89.9} \\
        \rowcolor{gray!20}\Checkmark &   \Checkmark  &  \Checkmark  & \Checkmark & \textbf{64.9} & \textbf{67.7} & 89.5  \\
        \bottomrule 
        \end{tabular}
        \label{tab:ablation_all}        
\end{table}
\subsubsection{Weighting Factor of PMM}
    \begin{table}[t]
\caption{Effect of the weighting factor $\lambda_{P}$ used for pixel motion matching. The results show that incorporating PMM improves association accuracy across a wide range of values.}  
        \centering
              
        \begin{tabular}{c |c c c c c}        
        \toprule
         $\lambda_{P}$ & HOTA↑ & IDF1↑  & AssA↑ & DetA↑ & AssR↑ \\
        \midrule     
                             0.1  & 53.1 & 53.0 & 36.6 & \textbf{77.4} & 42.0\\
                             \rowcolor{gray!20}0.2  &\textbf{53.3} & 53.7 & \textbf{36.8} & \textbf{77.4} & \textbf{42.4}\\
                             0.3  &\textbf{53.3} & 53.7 & \textbf{36.8} & \textbf{77.4} & 42.3\\
                             0.4  & 53.2 & \textbf{53.9} & 36.7 & 77.3 & 42.3\\        
                             0.5  & 52.8 & 53.6 & 36.2 & 77.1 & 42.1\\        
                             0.6  & 52.7 & 53.4 & 36.1 & 77.2 & 42.0\\    
        \bottomrule
        \end{tabular}
        \label{tab:ablation_pmm}
    \end{table}

To effectively fuse instantaneous pixel-level cues with KF-predicted instance-level cues in the association matrix, we study the effect of the PMM weighting hyperparameter $\lambda_{P}$, as summarized in Table~\ref{tab:ablation_pmm}.
Results show that incorporating PMM consistently improves performance across most $\lambda_{P}$ values, indicating that masked optical flow effectively captures target motion.
The best results in HOTA, AssA, DetA, and AssR, along with near-optimal IDF1, are obtained when $\lambda_{P} = 0.2$, which we adopt in our implementation.

\subsubsection{Weighting Factor of CDM}
    \begin{table}[t]
\caption{Effect of the weighting factor $\lambda_{C}$ used for the CDM module. The results show that an appropriate weighting improves association accuracy and overall tracking performance.}  
        \centering
        
        \begin{tabular}{c |c c c c c }        
        \toprule
         $\lambda_{C}$ & HOTA↑ & IDF1↑  & AssA↑ & DetA↑ & AssR↑ \\
        \midrule
        0.0 (BYTE-only)  & 52.9 & 52.4 & 35.9 & 78.2 & 42.0\\
        0.1              & 53.0 & 52.6 & 36.1 & 78.2 & 42.2\\
        0.2              & 53.1 & 52.5 & 36.2 & 78.2 & 42.2\\
        0.3              & 53.1 & 52.8 & 36.2 & 78.2 & 42.3\\
        0.4              & 53.0 & 52.4 & 36.1 & 78.2 & 42.1\\        
        0.5              & 53.1 & 52.6 & 36.2 & 78.2 & 42.2\\        
        0.6              & 53.1 & 52.5 & 36.2 & 78.1 & 42.1\\        
        0.7              & 53.1 & 52.7 & 36.2 & 78.2 & 42.3\\        
        0.8              & 53.3 & 53.0 & 36.5 & 78.0 & 42.7\\        
        0.9              & 53.3 & 53.0 & 36.6 & 78.1 & 42.8\\        
        \rowcolor{gray!20}1.0              & \textbf{53.5} & \textbf{53.3} & \textbf{36.7} & \textbf{78.3 }& \textbf{42.9}\\        
        \bottomrule
        \end{tabular}
        \label{tab:ablation_cdm}
    \end{table}

We also study the effect of the CDM weighting factor $\lambda_{C}$, as summarized in Table~\ref{tab:ablation_cdm}.
Since CDM targets low-confidence detections, we introduce an additional BYTE association step~\cite{zhang2022bytetrack} for these detections when CDM is enabled. This extra matching round relies solely on IoU-based association.
When $\lambda_{C}=0$ (i.e., the BYTE-only configuration), the additional association already improves performance by utilizing more detections.
As $\lambda_{C}$ increases, the tracker further improves detection utilization and association accuracy, yielding gains of +0.6 HOTA, +0.9 IDF1, +0.8 AssA, +0.1 DetA, and +0.9 AssR compared with the BYTE-only setting.
These results indicate that CDM effectively captures characteristics of low-confidence targets, enabling more flexible and reliable association.

\subsubsection{Update Threshold in DBC}

We analyze the effect of the update threshold $\tau_{D}$.
A small threshold may include samples unrelated to true long-tailed motion events, while a large value may lead to insufficient samples for reliable modeling.
As shown in Table~\ref{tab:ablation_dbc_thresh}, the best performance and association accuracy are achieved when $\tau_{D}=0.1$, which we adopt in our implementation.

\subsubsection{Effectiveness of DBC}
\begin{table}[t]
\caption{Effect of the update threshold $\tau_{D}$ in the DBC module. 
The threshold controls when distribution-based correction is activated for potential long-tailed motion events. 
Best overall tracking and association performance is achieved when $\tau_{D}=0.1$.}
        \centering
         
        \begin{tabular}{c |c c c c c}        
        \toprule
         $\tau_{D}$ & HOTA↑ & IDF1↑  & AssA↑ & DetA↑ & AssR↑\\
        \midrule
        0.05             & 52.4 & 52.2 & 35.6 & 77.2 & 40.4\\
        \rowcolor{gray!20}0.1              & \textbf{52.6} & \textbf{52.5} &\textbf{35.9} & 77.4 & \textbf{40.7}\\
        0.2              & 52.3 & 52.0 & 35.4 & 77.4 & 40.3\\
        0.3              & 51.8 & 51.1 & 34.7 & \textbf{77.5} & 39.5\\
        0.4              & 52.1 & 51.7 & 35.2 & \textbf{77.5} & 40.0\\        
        0.5              & 52.4 & 52.1 & 35.5 & 77.4 & 40.5\\        
        0.6              & 52.2 & 51.9 & 35.3 & 77.4 & 40.2\\       
        \bottomrule
        \end{tabular}
        \label{tab:ablation_dbc_thresh}
\end{table}

\begin{table}[t]
\caption{Impact of different correction strategies. 
Our DBC selectively captures long-tailed motion events and triggers correction only when necessary, leading to improved association accuracy compared to applying correction to all pairs.}   
        \centering
        
        \begin{tabular}{c |c c c c c}        
        \toprule
         Method & HOTA↑ & IDF1↑  & AssA↑ & DetA↑ & AssR↑\\
        \midrule
        Correct All     & 51.7 & 50.6 & 34.9 & 77.1 & 39.6\\
         \rowcolor{gray!20}DBC             & \textbf{52.6} & \textbf{52.5} & \textbf{35.9} & \textbf{77.4} & \textbf{40.7}\\
        \bottomrule
        \end{tabular}
        \label{tab:ablation_dbc}
\end{table}

Table~\ref{tab:ablation_dbc} evaluates the effectiveness of our DBC correction strategy for IoU computation.
Naively applying IoU correction to all detection-track pairs degrades performance because long-tailed motion occurs only sporadically during tracking.
By detecting abnormal motion and activating correction through the distribution-based strategy, DBC mitigates motion-matching degradation caused by bounding-box deformation and improves association robustness.

\subsubsection{Effectiveness of Cluster-Aware Update Strategy}

\begin{table}[t]
\caption{Impact of the cluster-aware selective update strategy. 
CA-ReID reduces noise in the maintained trajectory appearance features by selectively updating embeddings from reliable detections, leading to improved association accuracy and overall tracking performance.}
\centering
\begin{tabular}{c |c c c c c}
\toprule
Method & HOTA↑ & IDF1↑ & AssA↑ & DetA↑ & AssR↑\\
\midrule
Simple ReID Update & 59.6 & 60.5 & 45.2 & 79.0 & 50.9\\
\rowcolor{gray!20}CA-ReID Update & \textbf{60.6} & \textbf{61.1} & \textbf{46.9} & \textbf{79.1} & \textbf{52.2}\\
\bottomrule
\end{tabular}
\label{tab:ablation_cau}
\end{table}

Table~\ref{tab:ablation_cau} evaluates the cluster-aware update strategy.
Updating trajectory appearance features using all matched embeddings (Simple ReID Update) degrades tracking and association accuracy due to low-quality embeddings.
In contrast, the cluster-aware strategy detects potential occlusions and selectively updates features, reducing noise accumulation and producing more discriminative trajectory embeddings.

\subsubsection{Comparison with Other Optical Flow Method}

For optical flow, we adopt SEA-RAFT, which offers a better trade-off between cross-dataset generalization and runtime efficiency.
Keeping all other settings identical, we compare Farneback with SEA-RAFT in Table~\ref{tab:Farneback_searaft}.
SEA-RAFT consistently improves all metrics, indicating that accurate optical flow estimation is crucial for exploiting pixel-level motion cues.

\begin{table}[t]
\caption{Comparison between Farneback and SEA-RAFT for optical flow estimation. 
SEA-RAFT provides more accurate motion estimation than the traditional Farneback method, leading to consistent improvements across all tracking metrics.}
\centering
\small
\begin{tabular}{c |c c c}
\toprule
Method & HOTA↑ & IDF1↑ & MOTA↑ \\
\midrule
Baseline & 52.2 & 51.9 & 87.3\\
Farneback & 53.8 & 52.7 & 87.2\\
\rowcolor{gray!20}SEA-RAFT & \textbf{55.8} & \textbf{56.2} & \textbf{88.9}\\
\bottomrule
\end{tabular}
\label{tab:Farneback_searaft}
\end{table}

\subsubsection{Comparison with Heuristic Filtering Strategy}

As shown in Table~\ref{tab:pdf_sam}, removing pixel filtering (i.e., using all bounding-box pixels) causes a noticeable performance drop.
This occurs because optical flow extracted from the entire bounding box inevitably includes background pixels and pixels belonging to nearby objects.
These noisy pixels introduce inconsistent motion signals that degrade the reliability of pixel-level motion cues.

To further evaluate the benefit of SAM, we design a heuristic alternative called Pixel Denoising Filtering (PDF).
PDF filters noisy pixels through two mechanisms.

\textit{i) Background Noise Filtering.}
Assuming background pixels exhibit minimal motion, we compute the optical flow magnitude for each pixel and apply min-max normalization within the target region.
Pixels with normalized flow magnitude below threshold $\tau_m$ are removed:
\begin{equation}
\tilde{M}_t = \left\{ (x, y) \big| 
\frac{m(p) - \min(m_t)}{\max(m_t) - \min(m_t)} \geq \tau_m
\right\}.
\end{equation}

\textit{ii) Interference from Adjacent Targets.}
We further apply a center-cross mask to suppress pixels near box boundaries where overlaps are likely:
\begin{equation}
\tilde{C}_t =
\left\{ (x, y) \big|
|x-x_c|\le \alpha W \vee |y-y_c|\le \alpha H
\right\}.
\end{equation}

The final key pixel set used for PMM is
\begin{equation}
P_t^{PDF} = \tilde{M}_t \cap \tilde{C}_t.
\end{equation}

Table~\ref{tab:pdf_sam} compares PDF and SAM filtering.
Both improve performance over the baseline, while SAM achieves the best results without introducing extra hyperparameters.
Here, the best PDF configuration uses $\tau_m=0.7$ and $2\alpha=0.9$.

\begin{table}[t]
\caption{Comparison between heuristic PDF filtering and SAM-based pixel selection. 
Both strategies aim to remove noisy pixels introduced by background regions or adjacent targets. 
While PDF relies on motion saliency and spatial heuristics to filter pixels, SAM directly produces accurate target masks. 
The results show that SAM-based filtering yields more reliable pixel-level cues and improves tracking performance.}
\centering
\small
\begin{tabular}{c |c c c}
\toprule
Method & HOTA↑ & IDF1↑ & MOTA↑\\
\midrule
Baseline & 52.2 & 51.9 & 87.3\\
w/o Filtering & 52.0 & 52.0 & 87.3\\
PDF Filtering & 52.9 & 53.2 & \textbf{87.4}\\
\rowcolor{gray!20}SAM Filtering & \textbf{53.3} & \textbf{53.7} & \textbf{87.4}\\
\bottomrule
\end{tabular}
\label{tab:pdf_sam}
\end{table}


\subsection{Qualitative Results}

\begin{figure}[t]
\centering
\includegraphics[width=1.\linewidth]{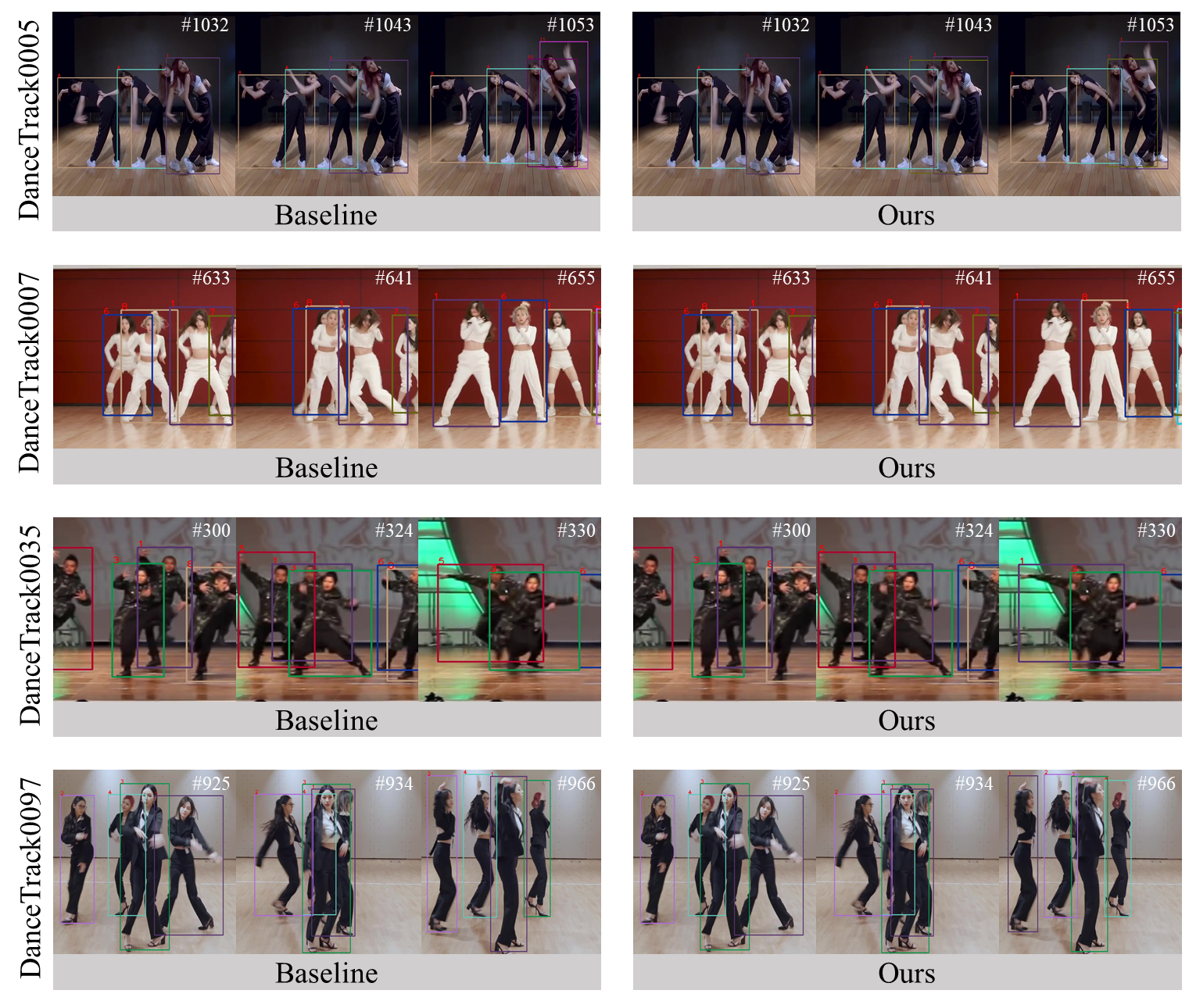}
\caption{Tracking comparison between the baseline tracker and SAMOFT. 
Under challenging scenarios such as occlusion, bounding-box deformation, and trajectory overlap, the baseline often produces identity switches, while SAMOFT maintains consistent identities by leveraging pixel-level motion cues and refined appearance updates.}
\label{fig:vis}
\end{figure}

Figure~\ref{fig:vis} shows qualitative comparisons under challenging scenarios.
The baseline suffers from identity switches under occlusion, severe bounding-box deformation, and trajectory overlap.
Examples include:
DanceTrack0005, where ID 1 switches after occlusion;
DanceTrack0007, where IDs 6 and 8 swap;
DanceTrack0035, where severe deformation causes ID 1 to switch to ID 5;
and DanceTrack0097, where heavy overlap causes cascading identity switches.
SAMOFT maintains consistent identities by leveraging reliable pixel-level motion cues and improved appearance updates.

\begin{figure}[t]
\centering
\vspace{-1.0em}
\includegraphics[width=1.\linewidth]{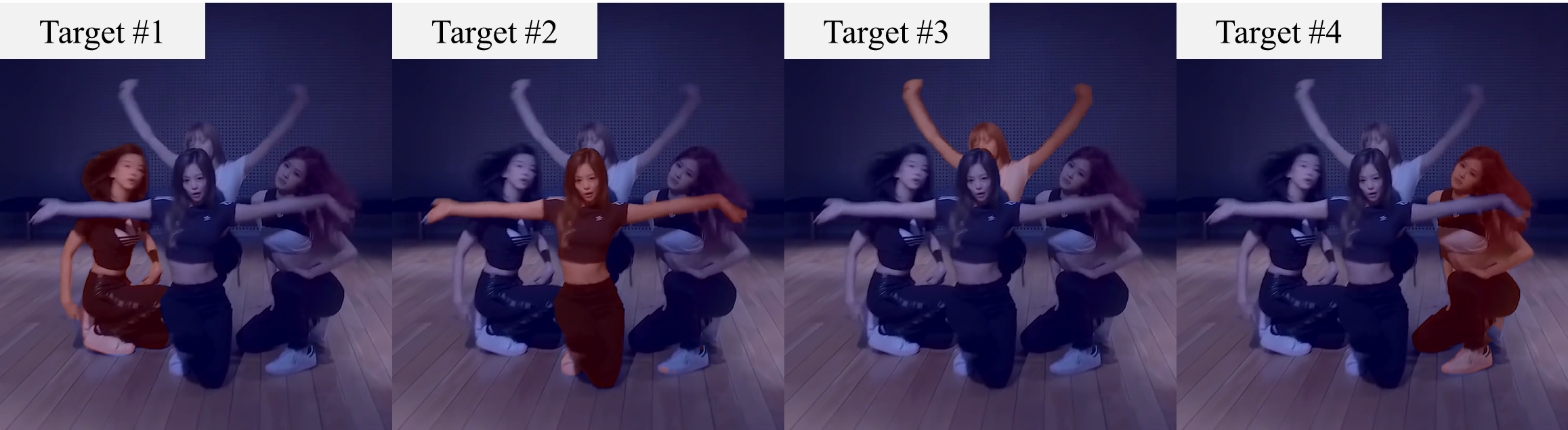}
\vspace{-0.8em}
\caption{SAM segmentation under partial occlusion. Despite significant overlap between targets, SAM is able to produce accurate masks that isolate individual objects. These masks provide reliable pixel-level cues that support robust motion estimation and trajectory association in our framework.}
\label{fig:occlu_vis}
\vspace{-1.4em}
\end{figure}

A potential concern is that near-complete occlusion may limit segmentation-based modules.
However, results in Tables~\ref{tab:benchmark} and \ref{tab:benchmark_mot20} show that our method still outperforms baselines on datasets with frequent occlusions.
Under severe occlusion, motion prediction from historical states remains useful, while under moderate occlusion SAM still provides reliable segmentation (Fig.~\ref{fig:occlu_vis}).
Long-term tracking under complete occlusion remains an open problem for future work.

\subsection{Limitations and Prospects}

\subsubsection{Computational Overhead}

SAMOFT introduces additional computation by integrating SAM segmentation and SEA-RAFT optical flow.
On the DanceTrack validation set, the baseline runs at 28.6 FPS.
Adding SEA-RAFT reduces speed to 23.8 FPS, while integrating SAM further reduces it to 14.1 FPS.

Despite this overhead, the improved performance justifies the use of pixel-level cues.
We expect this gap to narrow as lightweight segmentation and optical flow models continue to improve.

\subsubsection{Future Directions}

Our design follows a compact principle to explore pixel-level cues within SORT-style tracking-by-detection frameworks.
Future work may further improve pixel-level modeling by developing lightweight segmentation and optical flow models tailored for MOT, designing more advanced interaction mechanisms between motion and segmentation, or exploring learnable fusion strategies.

Ultimately, fully pixel-level cue-driven trackers may replace traditional bounding-box-based tracking-by-detection approaches.

\section{Conclusion}
\label{sec:conclusion}

In this work, we show that pixel-level cues derived from segmentation and optical flow can effectively improve the association robustness of MOT trackers. 
Instantaneous pixel motion complements Kalman filter-based motion prediction when historical observations become unreliable, while mask-based spatial cues provide more stable localization under partial occlusion. 
Furthermore, modeling the statistical distribution of optical flow magnitudes enables the detection of long-tailed motion patterns, allowing adaptive correction of trajectory states under complex motion conditions. 
In addition, awareness of target states reduces the effect of low-quality embeddings on maintained trajectory features.

Based on these insights, we propose SAMOFT, a robust tracking framework that achieves strong performance on multiple public benchmarks, validating the effectiveness of the proposed design. 
We hope this work encourages further exploration of integrating context-aware and pixel-level cues into MOT systems, potentially complementing or replacing traditional instance-level matching strategies and introducing statistical reasoning into online trajectory association.

{
    \bibliographystyle{IEEEtran}
    \bibliography{my_references}
}

\vfill

\end{document}